\documentclass[pdflatex,sn-mathphys-num]{sn-jnl}

\usepackage{graphicx}%
\usepackage{multirow}%
\usepackage{amsmath,amssymb,amsfonts}%
\usepackage{amsthm}%
\usepackage{mathrsfs}%
\usepackage[title]{appendix}%
\usepackage{xcolor}%
\usepackage{textcomp}%
\usepackage{manyfoot}%
\usepackage{booktabs}%
\usepackage{algorithm}%
\usepackage{algorithmicx}%
\usepackage{algpseudocode}%
\usepackage{listings}%

\usepackage{lineno}%

\usepackage{float}%

\theoremstyle{thmstyleone}%
\theoremstyle{thmstyletwo}%

\theoremstyle{thmstylethree}%

\begin{document}

\title[Article Title]{Chaotic CNN for Limited Data Image Classification}


\author[1]{\fnm{Anusree} \sur{M}}\email{1999anusreem@gmail.com}

\author[1]{\fnm{Akhila} \sur{Henry}}\email{akhilahenryu@am.amrita.edu}

\author*[1]{\fnm{Pramod} \sur{P Nair}}\email{pramodpn@am.amrita.edu}

\affil*[1]{\orgdiv{Department of Mathematics}, \orgname{Amrita Vishwa Vidyapeetham}, \orgaddress{\city{Amritapuri}, \country{India}}}


\abstract{Convolutional neural networks (CNNs) often exhibit poor generalization in limited training data scenarios due to overfitting and insufficient feature diversity. In this work, a simple and effective chaos-based feature transformation is proposed to enhance CNN performance without increasing model complexity. The method applies nonlinear transformations using logistic, skew tent, and sine maps to normalized feature vectors before the classification layer, thereby reshaping the feature space and improving class separability. The approach is evaluated on grayscale datasets (MNIST and Fashion-MNIST) and an RGB dataset (CIFAR-10) using CNN architectures of varying depth under limited data conditions. The results show consistent improvement over the standalone (SA) CNN across all datasets. Notably, a maximum performance gain of 5.43\% is achieved on MNIST using the skew tent map with a 3-layer CNN at 40 samples per class. A higher gain of 9.11\% is observed on Fashion-MNIST using the sine map with a 3-layer CNN at 50 samples per class.  Additionally, a strong gain of 7.47\% is obtained on CIFAR-10 using the skew tent map at 200 samples per class. The consistent improvements across different chaotic maps indicate that the performance gain is driven by the shared nonlinear and dynamical properties of chaotic systems. The proposed method is computationally efficient, requires no additional trainable parameters, and can be easily integrated into existing CNN architectures, making it a practical solution for data-scarce image classification tasks.}




\maketitle
\section{Introduction}\label{sec1}

Deep learning models, especially convolutional neural networks (CNNs), have achieved strong performance in image classification tasks. However, these models typically require large amounts of labelled data for effective training. In many real-world applications, obtaining sufficient labelled data is difficult, expensive, or time-consuming. As a result, CNNs often suffer from overfitting and poor generalization in limited data scenarios. Several techniques have been proposed to address this issue. Recently, there has been growing interest in incorporating concepts from nonlinear dynamical systems into machine learning models (\cite{anusree2025understanding}). Chaotic systems are characterized by strong nonlinearity, bounded behavior, and high sensitivity to initial conditions. These properties can be useful for enhancing feature representation (\cite{harikrishnan2020neurochaos}) and introducing controlled perturbations (\cite{mizutani1998controlling}) in learning systems.

In this work, a simple yet effective approach is proposed to improve CNN performance under limited training data. A chaotic transformation based on the logistic map is applied to normalized feature vectors obtained from convolutional layers. This transformation is introduced just before the fully connected classification layer. The proposed method does not add significant computational complexity and can be easily integrated into existing CNN architectures. To further investigate the generality of the approach, additional chaotic maps are also considered. Experimental results on grayscale and color image datasets, using CNN models of varying depth, show consistent improvement in classification performance. These findings suggest that the observed gains are not specific to a single chaotic map, but are instead related to the underlying properties of chaotic transformations.


\section{Literature Review}
\label{sec:literature}

Deep learning models require large labelled datasets for effective training. In many real-world applications, such data is not available. This leads to overfitting and poor generalization. The model tends to memorize training samples instead of learning meaningful patterns. Several approaches have been proposed to address this issue. Data augmentation (\cite{shorten2019survey}) increases the size of the dataset by applying transformations to input samples. Regularization methods such as dropout (\cite{wu2015towards}) and weight decay (\cite{park2019bayesian}) reduce model complexity. Semi-supervised learning methods use both labelled and unlabelled data to improve performance (\cite{liu2017semi}). Transfer learning reuses knowledge from pre-trained models (\cite{gupta2022deep}). However, these methods have certain limitations. Data augmentation may not capture true data variability. Regularization methods may reduce model capacity. Semi-supervised methods depend on the quality of pseudo-labels. Transfer learning requires suitable pre-trained models and domain similarity. Therefore, improving performance under limited data conditions remains an open challenge.

\subsection{CNN Architectures for Image Classification in Low Data Scenarios}

Convolutional neural networks are widely used for image classification tasks. They learn hierarchical features through convolution and pooling operations. Standard CNN architectures perform well when large datasets are available. For low data scenarios, several variations of CNNs have been proposed. Shallow networks (\cite{gao2018sd}) are used to reduce overfitting. Pre-trained deep networks (\cite{hasan2017application}) are fine-tuned using small datasets. Few-shot (\cite{wang2019few}) and meta-learning (\cite{wang2019meta}) approaches aim to learn from very few samples. Regularized CNNs include dropout and batch normalization layers to improve generalization.

Despite these advancements, challenges still exist. Shallow networks may fail to capture complex patterns. Fine-tuning may lead to overfitting if the dataset is very small. Meta-learning methods are often complex and computationally expensive. Regularization techniques may not fully prevent overfitting. Therefore, there is a need for simple and effective methods that improve feature representation without increasing model complexity.

\subsection{Chaotic Neural Networks}

Chaotic neural networks combine principles of chaos theory and neural computation (\cite{anusree2025understanding}). They introduce nonlinear dynamics into learning systems. These networks exhibit properties such as sensitivity to initial conditions, bounded behavior, and complex trajectories (\cite{henry2025simplified}). Previous studies have shown that chaotic dynamics can enhance learning ability. Chaotic systems can explore complex solution spaces efficiently (\cite{RemyaAjai2025}). They have been applied in pattern recognition (\cite{crook2001novel}), optimization (\cite{chen1995chaotic}), and time-series prediction (\cite{fukuda2021analysis}) tasks. Models such as Echo State Networks (\cite{jaeger2002adaptive}) use chaotic reservoirs for processing temporal data. Neurochaos learning frameworks use chaotic maps to extract features from input data (\cite{harikrishnan2020neurochaos}).

Despite their advantages, chaotic neural networks also have limitations. Controlling chaotic behavior is challenging. The performance is sensitive to parameter settings. Training can be computationally complex in some architectures. In addition, most existing works focus on standalone chaotic models rather than integrating chaos into standard deep learning pipelines.

\subsection{Chaotic Maps}

In this work, three chaotic maps are considered, namely the logistic map (\cite{RemyaAjai2023a}), the skew tent map (\cite{Balakrishnan2019a}), and the sine($\pi*x$) map \cite{Henry2025a}. These maps are simple one-dimensional nonlinear systems. They are widely used in chaos-based learning methods. All three maps operate on values in the interval $[0,1]$, which makes them suitable for feature transformation after normalization.

The \textbf{logistic map} is defined as:
\begin{equation}
	x_{n+1} = r x_n (1 - x_n)
\end{equation}
where $x \in [0,1]$ and $r \in (0,4]$. For values of $r$ close to 4, the system exhibits chaotic behavior. The map is highly nonlinear and sensitive to initial conditions. It produces bounded but irregular outputs. These properties help in increasing feature diversity and improving class separability.

The \textbf{skew tent map} is defined as:
\begin{equation}
	x_{n+1} =
	\begin{cases}
		\frac{x_n}{p}, & 0 \leq x_n < p \\
		\frac{1 - x_n}{1 - p}, & p \leq x_n \leq 1
	\end{cases}
\end{equation}
where $p \in (0,1)$ is a control parameter. This map is piecewise linear but exhibits chaotic dynamics. It has uniform distribution properties and strong mixing behavior. These characteristics help in redistributing feature values effectively across the interval.

The \textbf{sine map} is defined as:
\begin{equation}
	x_{n+1} = \sin(\pi x_n)
\end{equation}
This map is smooth and nonlinear. It generates complex trajectories within the bounded interval. It also shows sensitivity to initial values. The smooth nature of the sine map provides a different type of nonlinear transformation compared to the logistic and tent maps.

All three maps share common properties. They are nonlinear, bounded, and sensitive to initial conditions. These properties make them suitable for feature transformation in neural networks. When applied to normalized CNN features, they introduce controlled perturbations in the feature space. This helps in improving feature representation. Recent studies in chaos-based learning have shown that such transformations can enhance classification performance. In particular, improvements have been observed in low data (\cite{anusree8self}) and imbalanced data (\cite{anusree2024hypothetical})scenarios. The chaotic maps help in increasing feature variability without adding trainable parameters. This reduces overfitting and improves generalization. In the context of CNN-based image classification, these maps can be easily integrated as a transformation layer. Since CNN features can be normalized to $[0,1]$, the maps can be applied directly in an element-wise manner. The transformation does not change the dimensionality of the feature vector. It also introduces negligible computational cost.


\section{Proposed Methodology}
\label{sec:methodology}

In this work, a simple modification to the standard convolutional neural network (CNN) architecture is proposed to improve classification performance under limited training data conditions. The method introduces a chaotic transformation in the feature space, motivated by the nonlinear and dynamical properties of chaotic maps discussed in the previous section. These maps exhibit bounded behavior, strong nonlinearity, and sensitivity to initial conditions, which makes them suitable for enhancing feature representations.

The input image is first passed through a CNN consisting of convolutional and pooling layers. These layers extract hierarchical features from the input data. The output of the CNN is a feature vector denoted as $f = \Phi(x; \theta_c)$, where $\Phi(\cdot)$ represents the feature extraction process. This feature vector captures important spatial and structural information from the input image. Before applying the chaotic transformation, the feature vector is normalized to ensure that all values lie within the interval $[0,1]$. This step is necessary because the chaotic maps operate on bounded inputs. The normalized feature vector is denoted as $\tilde{f} = \mathcal{N}(f)$.

Following normalization, a chaotic transformation is applied to the feature vector. In this study, three different chaotic maps are considered, namely the logistic map, the skew tent map, and the sine map. These maps are applied element-wise to the normalized feature vector. For the logistic map, the transformation is given by $f^{*} = r \tilde{f} (1 - \tilde{f})$, where $r$ is a control parameter in the chaotic region. Here we have fixed $r=4$ for maximum chaos. For the skew tent map, the transformation is defined in a piecewise manner using a parameter $p \in (0,1)$. Here we have fixed $p=0.499$ for maximum entropy. For the sine map, the transformation is given by $f^{*} = \sin(\pi \tilde{f})$. These transformations preserve the dimensionality of the feature vector while introducing nonlinear perturbations. This helps in reshaping the feature space and improving class separability.

The transformed feature vector $f^{*}$ is then passed to a fully connected layer for classification. A softmax function is applied to obtain the final class probabilities, given by $\hat{y} = \text{softmax}(W f^{*} + b)$. The predicted label corresponds to the class with the highest probability. The proposed method is evaluated on three benchmark image datasets, namely MNIST, Fashion-MNIST, and CIFAR-10. The MNIST and Fashion-MNIST datasets consist of grayscale images with 10 classes and a total of 70,000 samples each. The CIFAR-10 dataset consists of RGB images with 10 classes and a total of 60,000 samples. All datasets have a balanced class distribution. The details of the datasets are given in Table\ref{tab:datasets}

\begin{table}[h]
	\centering
	\caption{Details of datasets used in this study}
\label{tab:datasets}
	\begin{tabular}{lccc}
		\hline
		Dataset & No. of Samples & No. of Classes & Class Distribution Ratio \\
		\hline
		MNIST & 70000 & 10 & Balanced \\
		Fashion-MNIST & 70000 & 10 & Balanced \\
		CIFAR-10 & 60000 & 10 & Balanced \\
		\hline
	\end{tabular}
\end{table}

To simulate limited data scenarios, only a small number of samples per class are used for training. For the grayscale datasets, experiments are conducted using CNN models with 2 and 3 convolutional layers. For each model, 40, 50, and 60 samples per class are considered. For the RGB dataset, a deeper CNN model with 5 convolutional layers is used. In this case, experiments are conducted with 100, 150, and 200 samples per class. The model is trained in an end-to-end manner using the categorical cross-entropy loss function. The parameters of the CNN and the classification layer are optimized using standard gradient-based methods. The chaotic transformation does not introduce additional trainable parameters, which makes the method computationally efficient. The performance is measured using macro F1-score.

The hyperparameters of the proposed CNN models were tuned using a grid search strategy combined with 5-fold stratified cross-validation to ensure robustness under limited data conditions. For the grayscale datasets, both the 2-layer and 3-layer CNN architectures were tuned over key parameters including the number of convolutional filters, kernel sizes, and the size of the fully connected layer. The 3-layer model additionally incorporates an extra convolutional block while following a similar search strategy. Each hyperparameter configuration was evaluated across multiple folds, resulting in a substantial number of training iterations to ensure reliable model selection. For the RGB dataset (CIFAR-10), a deeper 5-layer CNN architecture was employed, where the convolutional structure was kept fixed and tuning primarily focused on the classifier design and the integration of the chaotic transformation layer. The model was trained for a fixed number of epochs using the Adam optimizer. This systematic tuning approach ensures fair comparison and reliable selection of optimal configurations under data-scarce settings.

Overall, the proposed approach integrates chaotic maps as a feature transformation layer within the CNN framework. The method leverages the nonlinear and dynamical properties of chaotic systems to enhance feature representation. This leads to improved classification performance, especially in scenarios with limited training data.


\section{Results and Discussion}
\label{sec:results}

\subsection{Grayscale Image Datasets}

The results are shown in Tables \ref{tab:mnist_results} and \ref{tab:fmnist_results}. The reported values correspond to macro F1-scores, which provide a balanced evaluation across all classes, especially under limited data conditions. The baseline CNN is denoted as SA. The chaotic maps used are Logistic (L), Skew Tent (ST), and Sine($\pi x$) (SP).

\begin{table}[h]
	\centering
	\caption{F1 scores of MNIST dataset}
	\label{tab:mnist_results}
	\begin{tabular}{lccccc}
		\hline
		Samples/Class & Model & SA & L & ST & SP \\
		\hline
		40 & 2 Conv & 0.8763 & 0.9027 & 0.9026 & 0.8871 \\
		50 & 2 Conv & 0.8914 & 0.9072 & 0.9150 & 0.9166 \\
		60 & 2 Conv & 0.9034 & 0.9224 & 0.9230 & 0.9199 \\
		\hline
		40 & 3 Conv & 0.8619 & 0.8654 & 0.9087 & 0.8829 \\
		50 & 3 Conv & 0.8837 & 0.9095 & 0.9014 & 0.9070 \\
		60 & 3 Conv & 0.9031 & 0.9066 & 0.9256 & 0.9022 \\
		\hline
	\end{tabular}
\end{table}

\begin{table}[h]
	\centering
	\caption{F1 scores of Fashion-MNIST dataset}
	\label{tab:fmnist_results}
	\begin{tabular}{lccccc}
		\hline
		Samples/Class & Model & SA & L & ST & SP \\
		\hline
		40 & 2 Conv & 0.7576 & 0.7928 & 0.7683 & 0.7683 \\
		50 & 2 Conv & 0.7629 & 0.7988 & 0.7614 & 0.7615 \\
		60 & 2 Conv & 0.7795 & 0.7969 & 0.7902 & 0.7902 \\
		\hline
		40 & 3 Conv & 0.7571 & 0.7663 & 0.7623 & 0.7644 \\
		50 & 3 Conv & 0.7210 & 0.7880 & 0.7751 & 0.7867 \\
		60 & 3 Conv & 0.7542 & 0.7747 & 0.7718 & 0.7827 \\
		\hline
	\end{tabular}
\end{table}

From the results, it is observed that the chaotic models improve the performance compared to SA in most cases. The improvement is more noticeable when the number of samples per class is small.

For the MNIST dataset, all three chaotic maps provide consistent improvement over the baseline. The skew tent map shows strong performance, especially for the 3-layer CNN model. The sine map also performs well for the 2-layer CNN. The logistic map provides stable improvement across all configurations. For the Fashion-MNIST dataset, the logistic map shows the most consistent improvement across different sample sizes. The sine map also provides good performance, especially for the 3-layer CNN model. The skew tent map improves performance in some cases, but the improvement is relatively smaller compared to the other maps.

\begin{figure}[h]
	\centering
	\includegraphics[width=0.45\textwidth]{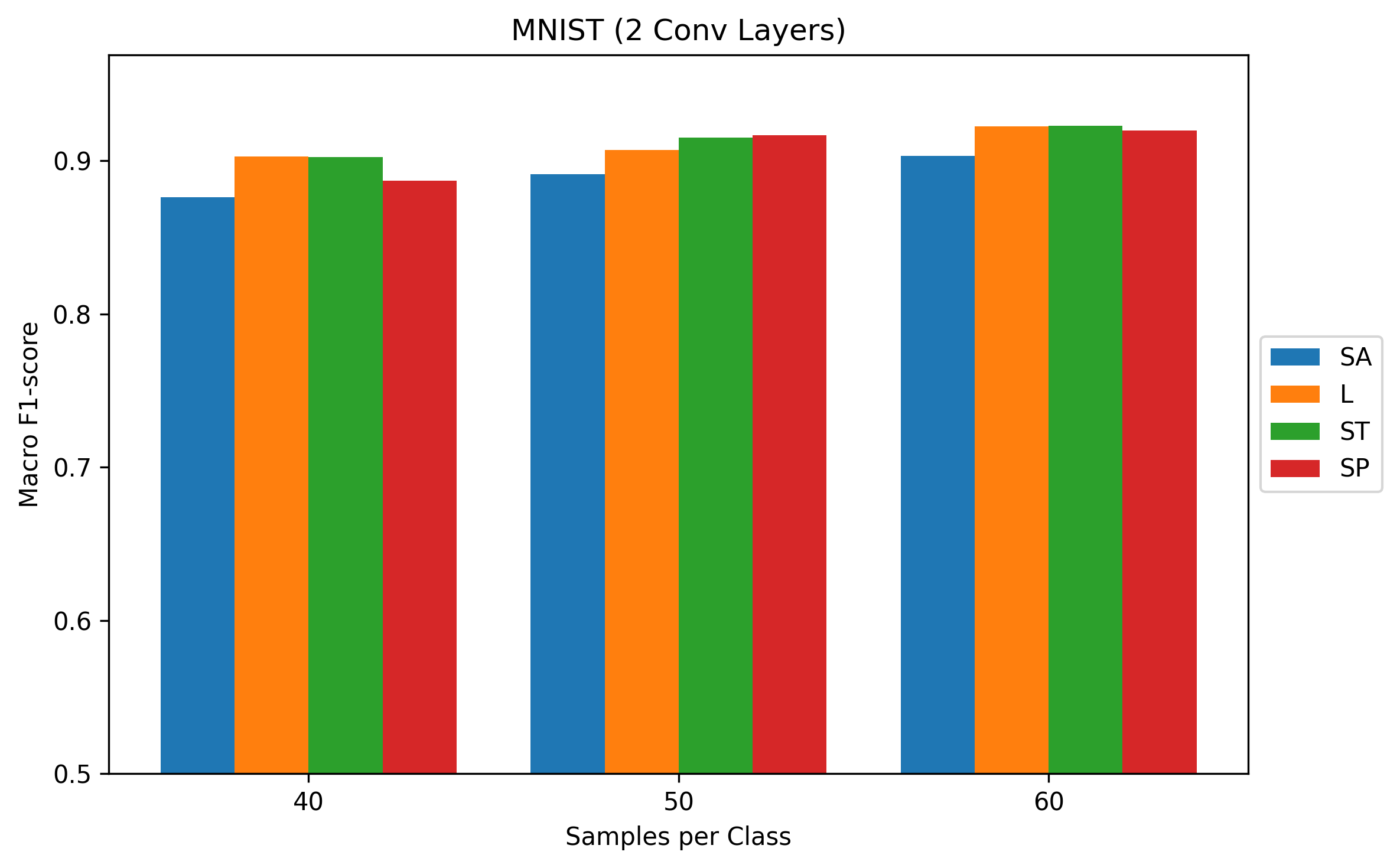}
	\includegraphics[width=0.45\textwidth]{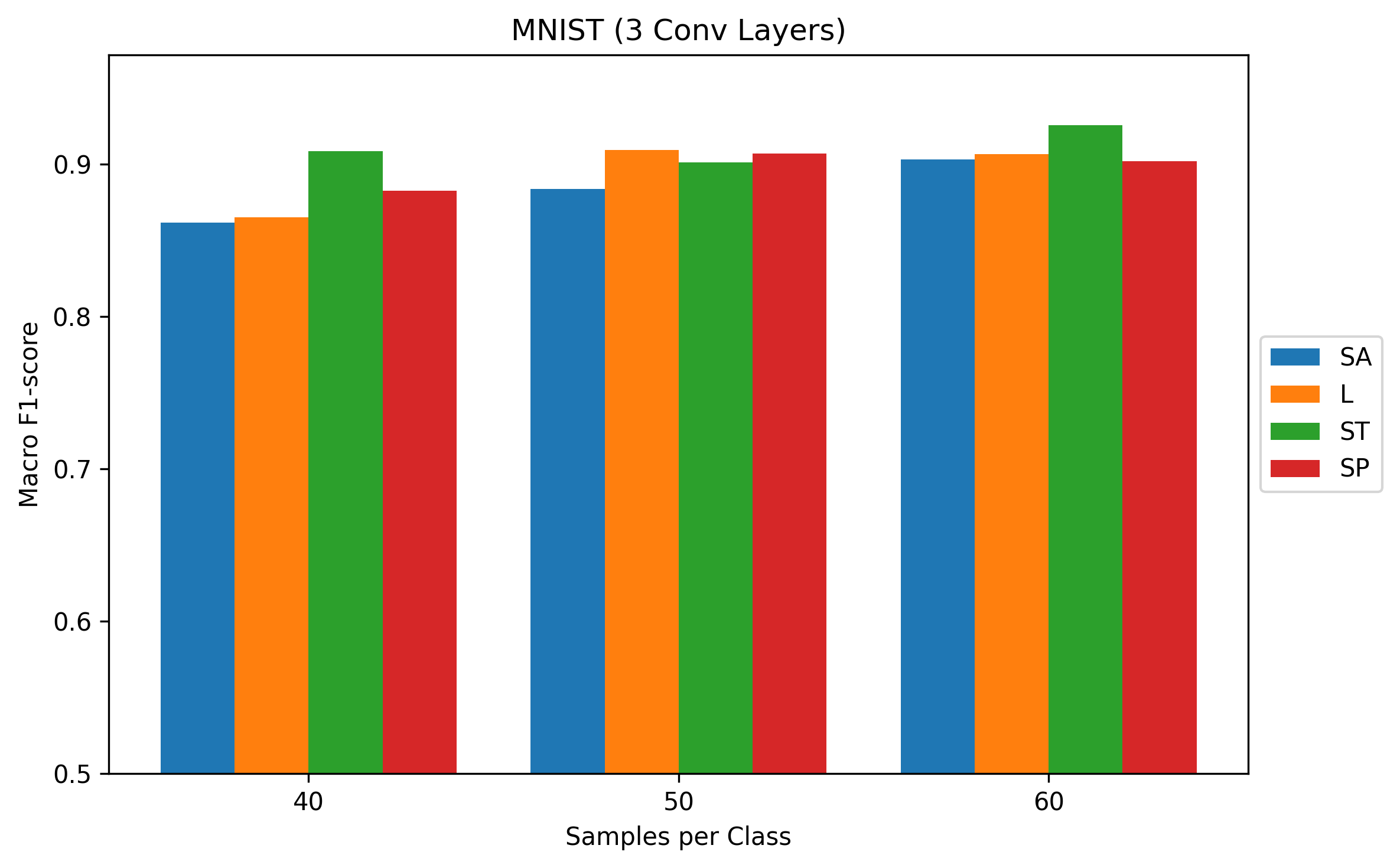}
	
	\vspace{0.2cm}
	
	\includegraphics[width=0.45\textwidth]{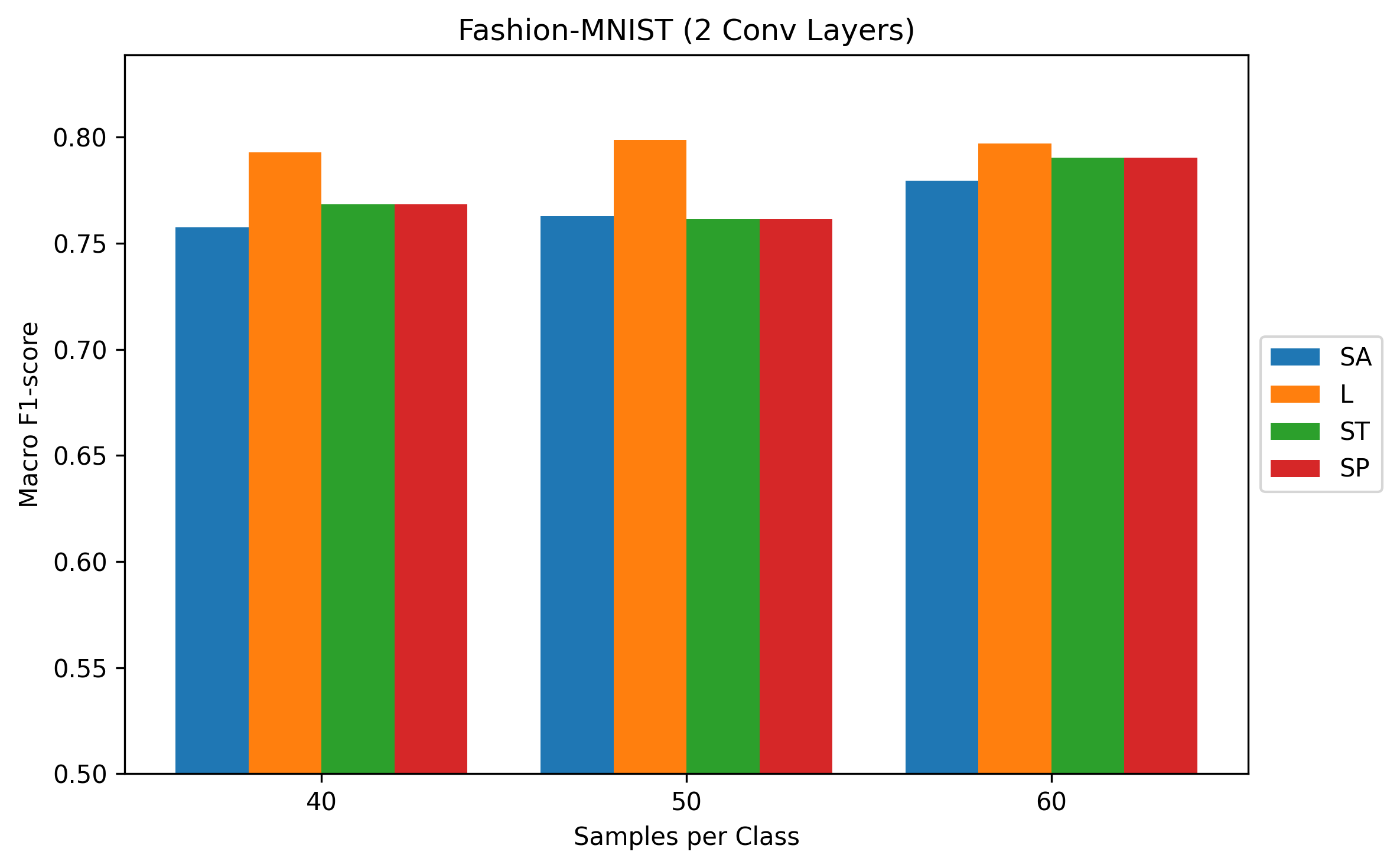}
	\includegraphics[width=0.45\textwidth]{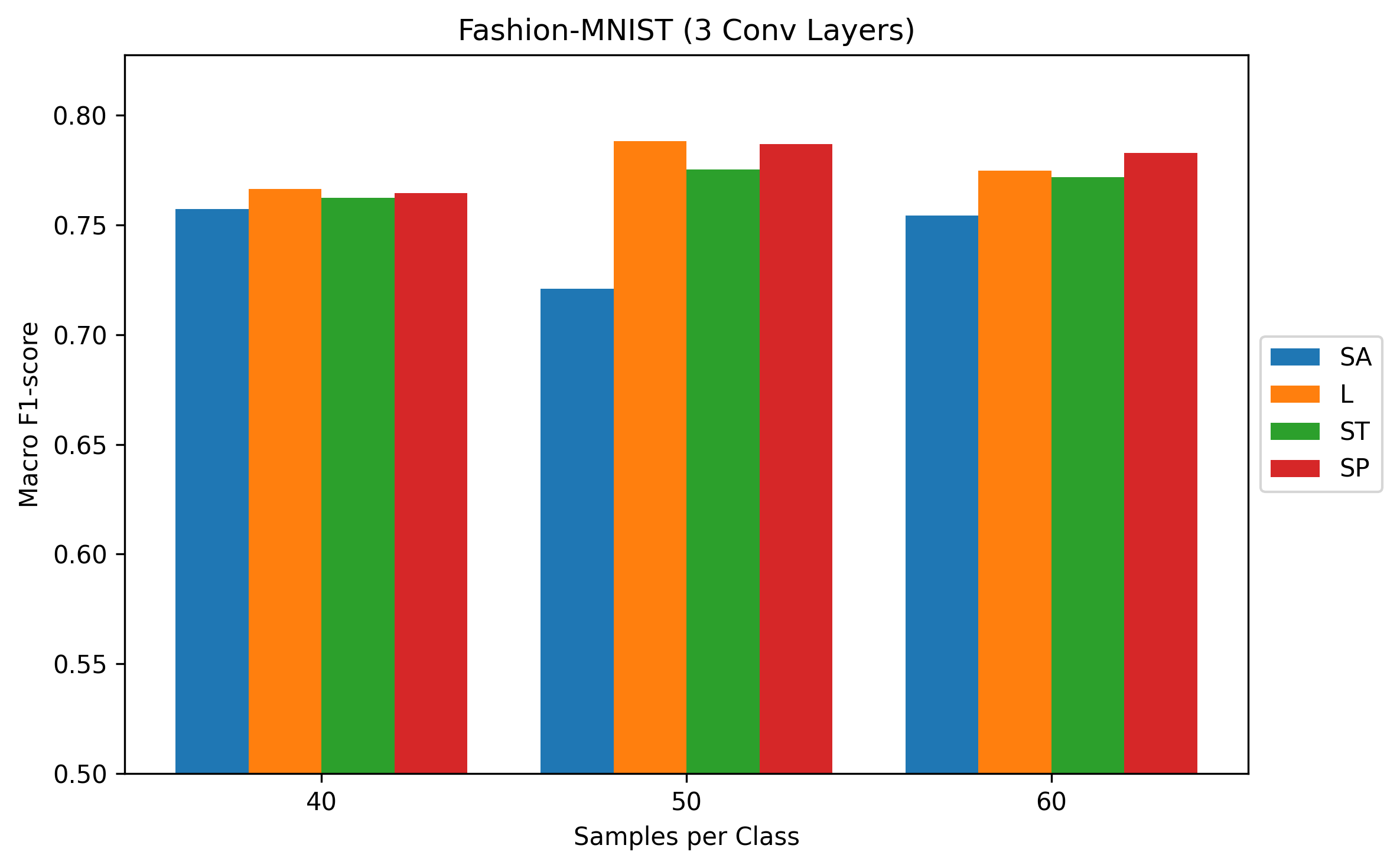}
	
	\caption{Comparison of macro F1-scores on grayscale datasets.}
	\label{fig:grayscale_results}
\end{figure}

It is also observed that the improvement is higher for lower sample sizes such as 40 samples per class. As the number of samples increases, the performance gap between the baseline and chaotic models reduces slightly. This indicates that the proposed chaotic transformation is more effective in highly limited data scenarios. Overall, the results show that integrating chaotic maps into CNN feature space improves classification performance for grayscale image datasets. The consistent improvement across different models and datasets suggests that the gain is due to the nonlinear and dynamical properties of chaotic transformations. Figure~\ref{fig:grayscale_results} shows the grouped bar plots comparing the macro F1-scores of the standalone (SA) CNN and the chaotic models for MNIST and Fashion-MNIST datasets. Each subplot corresponds to a specific CNN architecture (2-layer and 3-layer), and within each plot, three groups represent different numbers of samples per class, while individual bars denote the performance of SA, L, ST, and SP models. The plots clearly illustrate the consistent improvement achieved by the chaotic transformations across different sample sizes and CNN architectures, with certain maps showing stronger gains depending on the dataset and model depth.

\subsection{RGB Image Dataset}

The proposed method is also evaluated on the CIFAR-10 dataset. A 5-layer CNN model is used. The experiments are conducted with 100, 150, and 200 samples per class.

The results are shown in Table \ref{tab:cifar_results}.

\begin{table}[h]
	\centering
	\caption{F1 scores of CIFAR-10 dataset}
	\label{tab:cifar_results}
	\begin{tabular}{lccccc}
		\hline
		Samples/Class & SA & L & ST & SP \\
		\hline
		100 & 0.4050 & 0.4347 & 0.4088 & 0.4165 \\
		150 & 0.4477 & 0.4512 & 0.4638 & 0.4659 \\
		200 & 0.4513 & 0.4506 & 0.4850 & 0.4751 \\
		\hline
	\end{tabular}
\end{table}

From the results, it is observed that the chaotic models improve the performance compared to the SA CNN in most cases. The improvement is noticeable across all sample sizes.

For 100 samples per class, the logistic map provides the highest improvement, followed by the sine map. The skew tent map shows only a small improvement in this case. As the number of samples increases, the performance of the skew tent map improves significantly. For 150 and 200 samples per class, the skew tent map achieves the highest performance among all models. The sine map also shows strong and consistent improvement across all configurations. The logistic map provides improvement in most cases, but the gain is relatively smaller compared to the other maps. It is also observed that the improvement becomes more stable as the number of samples increases. This indicates that chaotic transformations are effective not only in very low data scenarios but also in moderately limited data conditions.

\begin{figure}[h]
	\centering
	\includegraphics[width=0.6\textwidth]{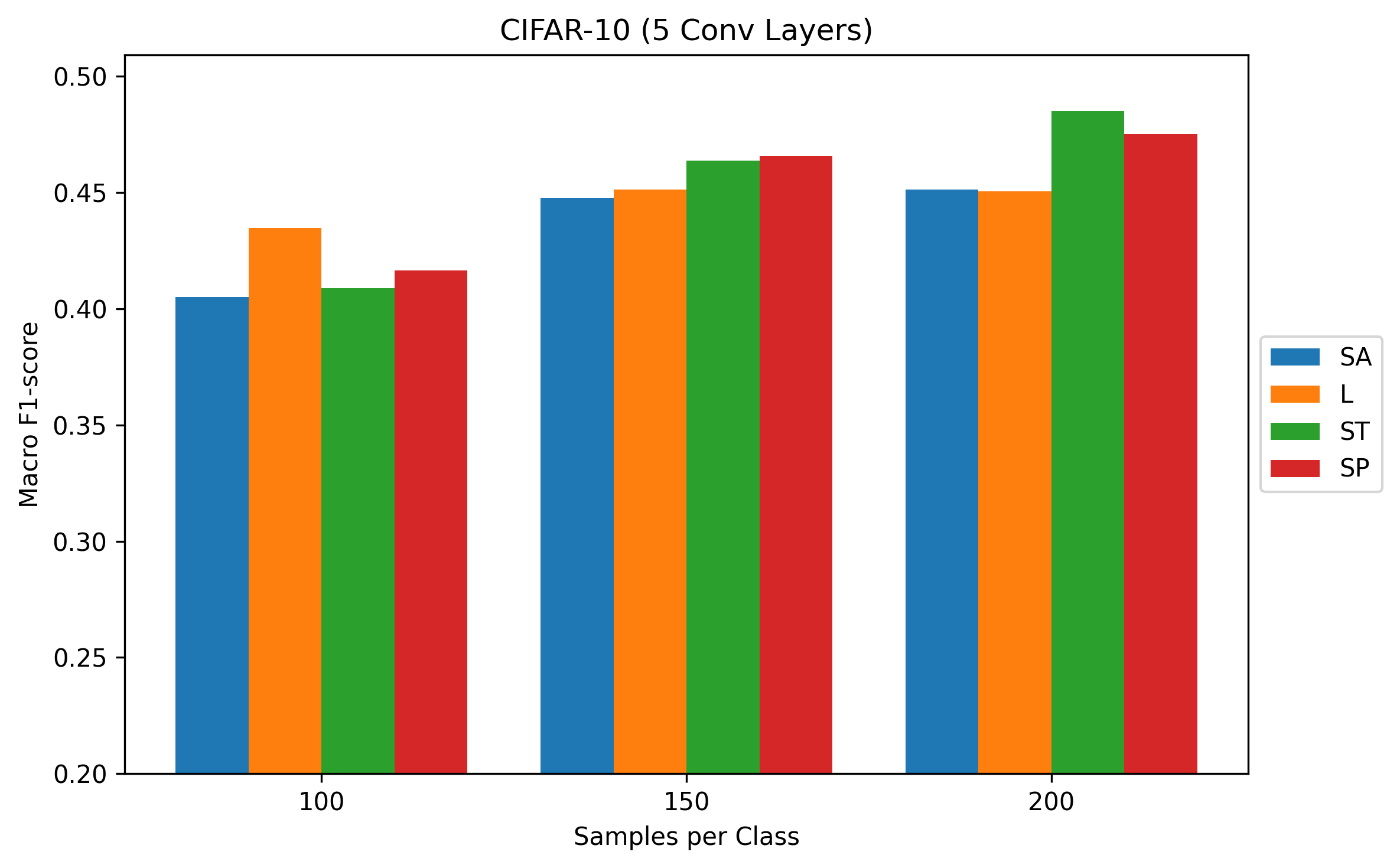}
	\caption{Comparison of macro F1-scores on CIFAR-10 dataset.}
	\label{fig:cifar_results}
\end{figure}

Overall, the results demonstrate that the proposed chaotic feature transformation enhances CNN performance for RGB image classification. The consistent improvement across different chaotic maps confirms the effectiveness of nonlinear feature transformation in improving representation and generalisation. Figure~\ref{fig:cifar_results} presents the grouped bar plot comparing the macro F1-scores of the SA CNN and the chaotic models on the CIFAR-10 dataset. The figure highlights the improvement achieved by the chaotic transformations, with the skew tent and sine maps showing strong and consistent performance across different sample sizes, particularly as the number of training samples increases.

\subsection{Performance Gain Analysis}

In this subsection, the performance improvement of the proposed chaotic models over the SA CNN is analysed using macro F1-score. The performance gain is computed in percentage as:

\begin{equation}
	\text{Gain (\%)} = \frac{F1_{\text{chaos}} - F1_{\text{SA}}}{F1_{\text{SA}}} \times 100
\end{equation}

where $F1_{\text{chaos}}$ represents the macro F1-score of the chaotic model and $F1_{\text{SA}}$ represents the macro F1-score of the SA CNN.

Tables \ref{tab:gain_mnist}, \ref{tab:gain_fmnist}, and \ref{tab:gain_cifar} present the percentage improvement achieved by the chaotic models for different datasets. The gain (\%) is explicitly calculated with respect to the macro F1-score of the SA CNN, and the tables are structured based on varying samples per class and CNN architectures, with separate columns showing the improvement obtained using L, ST, and SP maps. This representation provides a clear comparison of how each chaotic transformation contributes to performance enhancement under different experimental settings.

\begin{table}[h]
	\centering
	\caption{Gain \% on MNIST dataset}
	\label{tab:gain_mnist}
	\begin{tabular}{lcccc}
		\hline
		Samples/Class & Model & L (\%) & ST (\%) & SP (\%) \\
		\hline
		40 & 2 Conv & 3.01 & 3.00 & 1.23 \\
		50 & 2 Conv & 1.77 & 2.65 & 2.83 \\
		60 & 2 Conv & 2.10 & 2.17 & 1.83 \\
		\hline
		40 & 3 Conv & 0.41 & 5.43 & 2.44 \\
		50 & 3 Conv & 2.92 & 2.00 & 2.64 \\
		60 & 3 Conv & 0.39 & 2.49 & - \\
		\hline
	\end{tabular}
\end{table}

\begin{table}[h]
	\centering
	\caption{Gain \% on Fashion-MNIST dataset}
	\label{tab:gain_fmnist}
	\begin{tabular}{lcccc}
		\hline
		Samples/Class & Model & L (\%) & ST (\%) & SP (\%) \\
		\hline
		40 & 2 Conv & 4.65 & 1.41 & 1.41 \\
		50 & 2 Conv & 4.71 & - & - \\
		60 & 2 Conv & 2.23 & 1.37 & 1.37 \\
		\hline
		40 & 3 Conv & 1.22 & 0.69 & 0.96 \\
		50 & 3 Conv & 9.29 & 7.50 & 9.11 \\
		60 & 3 Conv & 2.72 & 2.33 & 3.78 \\
		\hline
	\end{tabular}
\end{table}

\begin{table}[h]
	\centering
	\caption{Gain \% on CIFAR-10 dataset}
	\label{tab:gain_cifar}
	\begin{tabular}{lccc}
		\hline
		Samples/Class & L (\%) & ST (\%) & SP (\%) \\
		\hline
		100 & 7.33 & 0.94 & 2.84 \\
		150 & 0.78 & 3.60 & 4.07 \\
		200 & - & 7.47 & 5.27 \\
		\hline
	\end{tabular}
\end{table}

For the grayscale datasets, a clear improvement is observed across most settings. The gain is higher for lower sample sizes such as 40 samples per class. This indicates that the proposed method is highly effective in very limited data conditions. For the MNIST dataset, the skew tent map shows the highest performance gain in most cases. The logistic map also provides consistent improvement across all configurations. The sine map shows moderate improvement, especially for the 2-layer CNN model. For the 3-layer CNN, the improvement is slightly reduced in some cases, but the overall trend remains positive. For the Fashion-MNIST dataset, the logistic map shows strong and consistent gain across different sample sizes. The sine map also performs well, particularly for the 3-layer CNN model. The skew tent map provides smaller gains compared to the other maps, and in a few cases the improvement is minimal. However, the overall performance is still comparable to or better than the baseline.

For the RGB dataset (CIFAR-10), a significant performance gain is observed. The skew tent map shows the highest improvement, especially for 150 and 200 samples per class. The sine map also provides strong and stable gains across all sample sizes. The logistic map improves performance, but the gain is comparatively smaller. It is also observed that the percentage gain decreases as the number of samples per class increases. This behaviour indicates that the chaotic transformation is more beneficial in highly data-scarce scenarios. As more training data becomes available, the baseline CNN itself learns better representations, and the relative gain reduces.

Overall, the performance gain analysis confirms that the proposed chaotic feature transformation consistently improves CNN performance. The improvement is observed across different datasets, architectures, and chaotic maps. This suggests that the gain is due to the general nonlinear and dynamical properties of chaotic systems, rather than any specific map formulation.


\section{Conclusion}

In this work, a chaos-based feature transformation method is proposed to improve CNN performance under limited training data conditions. The method applies nonlinear transformations using logistic, skew tent, and sine maps to normalized CNN features before classification. The approach is evaluated on grayscale datasets (MNIST and Fashion-MNIST) and an RGB dataset (CIFAR-10) using CNN models of varying depth. The results show consistent improvement over the SA CNN across most configurations. The performance gain is more significant in low data scenarios, confirming the effectiveness of the proposed method in data-scarce conditions.

The performance gain analysis further highlights that different chaotic maps contribute differently across datasets. The skew tent map shows strong improvement for MNIST and CIFAR-10, while the logistic and sine maps provide consistent gains for Fashion-MNIST. The improvement decreases as the number of training samples increases, indicating that the method is most beneficial when data is limited. The proposed approach is simple, computationally efficient, and does not introduce additional trainable parameters. As a future direction, the explainability of the proposed chaotic CNN model can be investigated to better understand the impact of chaotic transformations on feature representation and decision-making. Overall, the results demonstrate that chaos-based feature transformation is an effective and practical technique for enhancing CNN generalization in limited data scenarios.

\bibliography{cCNN}

\end{document}